# GPMFS: Global Foundation and Personalized Optimization for Multi-Label Feature Selection

Yifan Cao, Zhilong Mi, Ziqiao Yin, Binghui Guo, and Jin Dong

*Abstract*—**As artificial intelligence methods are increasingly applied to complex task scenarios, high dimensional multi-label learning has emerged as a prominent research focus. At present, the curse of dimensionality remains one of the major bottlenecks in high-dimensional multi-label learning, which can be effectively addressed through multi-label feature selection methods. However, existing multi-label feature selection methods mostly focus on identifying global features shared across all labels, which overlooks personalized characteristics and specific requirements of individual labels. This global-only perspective may limit the ability to capture label-specific discriminative information, thereby affecting overall performance. In this paper, we propose a novel method called GPMFS (Global Foundation and Personalized Optimization for Multi-Label Feature Selection). GPMFS firstly identifies global features by exploiting label correlations, then adaptively supplements each label with a personalized subset of discriminative features using a threshold-controlled strategy. Experiments on multiple real-world datasets demonstrate that GPMFS achieves superior performance while maintaining strong interpretability and robustness. Furthermore, GPMFS provides insights into the label-specific strength across different multi-label datasets, thereby demonstrating the necessity and potential applicability of personalized feature selection approaches.**

This work was supported by the National Science and Technology Major Project, grant: 2021ZD0201302, the National Natural Science Foundation of China, grant: 12201025, Zhongguancun Laboratory, the Fundamental Research Funds for the Central Universities, and the Beijing Advanced Innovation Center for Future Blockchain and Privacy Computing.

(Corresponding authors: Ziqiao Yin; Binghui Guo and Jin Dong.)

Yifan Cao is with the School of Artificial Intelligence, Beihang University, Beijing 100191, China, and Zhongguancun Laboratory, Beijing 100094, China (email: by2342101@buaa.edu.cn).

Zhilong Mi, Ziqiao Yin, and Binghui Guo are with the School of Artificial Intelligence, Beihang University, Beijing 100191, China; the Key Laboratory of Mathematics, Informatics and Behavioral Semantics, Beihang University; and the Beijing Advanced Innovation Center for Future Blockchain and Privacy Computing, Beihang University, Beijing 100191, China (emails: mizhilong@buaa.edu.cn, yinziqiao@buaa.edu.cn, guobinghui@buaa.edu.cn, respectively). Additionally, Ziqiao Yin and Binghui Guo are affiliated with Zhongguancun Laboratory, Beijing 100094, China. Ziqiao Yin is also affiliated with the Hangzhou International Innovation Institute, Beihang University, Hangzhou 311115, China.

Jin Dong is with the Beijing Academy of Blockchain and Edge Computing, Beijing 100080, China (email: dongjin@baec.org.cn).

## I. INTRODUCTION

With the rapid advancement of information technology, the scale and complexity of data have grown exponentially, rendering traditional single-label classification methods inadequate for many real-world applications. For instance, in text classification, a single news article may simultaneously relate to multiple topics such as "entertainment" and "science" [1]; in medical diagnosis, a patient may suffer from multiple diseases, such as hypoalbuminemia and pneumonia [2]; and in image annotation, a photo may be associated with both "mountains" and "fields" [3]. These scenarios fall under the domain of Multi-Label Classification (MLC), where the core task is to assign a set of relevant category labels to each sample [4]. Unlike single-label classification tasks, multi-label classification (MLC) involves unique challenges, including modeling label correlations, handling high-dimensional feature spaces, and dealing with class imbalance [5][6]. Moreover, MLC distinguishes itself by enabling algorithms to exploit the co-occurrence patterns among labels in multi-label datasets (MLDs), which can significantly enhance predictive performance [7]. Accordingly, MLC has become a prominent and active area of research in the field of machine learning.

The performance of multi-label learning algorithms heavily depends on the quality of the input features, highlighting the importance of effective feature selection in multi-label classification (MLC). Raw datasets often contain numerous redundant or irrelevant features [8], which not only increase computational complexity but also introduce noise that can degrade model performance [9]. Additionally, multi-label datasets are typically characterized by extremely high dimensionality. Multi-label feature selection addresses this "curse of dimensionality" by extracting the most informative features from a high-dimensional space associated with a large-scale label set [10], thereby improving generalization and enhancing model interpretability through the discovery of latent relationships between features and labels [11][12].

Based on their integration with the classification process, existing multi-label feature selection techniques are generally categorized into three main types: filter methods, wrapper methods, and embedded methods [13]. Filter-based methods perform feature selection by evaluating the correlations among features, independently of the classification task. Wrapper-based methods, on the other hand, utilize the classification performance to guide the selection process, formulating it as a search for the optimal feature subset. Embedded methods integrate feature selection directly into the model training process, enabling efficient modeling of feature interactions and improving computational efficiency [14]. These advantages make embedded methods the focus of this study.

Despite advancements, existing embedded multi-label feature selection methods, such as SCMFS [15], SLMDS [16], and LCIFS [17], typically select a single feature subset for all labels when handling label correlations. However, this approach may obscure label-specific discriminative features and complicate the establishment of clear relationships between features and labels, thus reducing interpretability. To address these challenges, we propose a novel hybrid feature selection framework: Global Foundation and Personalized Optimization for Multi-Label Feature Selection. This framework retains a globally shared feature subset while dynamically supplementing each label with personalized features via an adaptive mechanism.

Our main contributions are summarized as follows:

We propose a novel multi-label feature selection method that first exploits label correlations through a relaxed label mechanism, and then incorporates a Pearson correlation matrix constraint to reduce feature redundancy.

Building on the global subset, we further introduce a personalized selection threshold q to identify label-specific informative features. These personalized features are combined with the global subset to construct the final label-relevant feature set for each label.

Extensive experiments on multiple real-world datasets demonstrate that our method significantly outperforms existing approaches in terms of classification performance.

## II. RELATED WORK

In this section, we provide an overview of embedded, filter-based, and wrapper-based multi-label feature selection methods.

### A. Filter methods

Filter methods operate independently of specific multi-label classification (MLC) algorithms by evaluating the importance of each feature using predefined metrics, and selecting features based on thresholds or rankings. These methods are typically efficient and scalable, making them suitable for high-dimensional multi-label data.

Among early contributions, Lee et al. [18] introduced FIMF, a fast multi-label feature selection approach that utilizes information-theoretic feature ranking. Building upon this direction, Zhang et al. [19] developed an effective GRRO-based approach grounded in mutual information, which simultaneously incorporates feature relevance, feature redundancy, and label correlation to improve selection performance. More recently, Zhang et al. [20] proposed a method that jointly considers mutual information and interaction weights to enhance the discriminative power of selected features. In addition, Dai et al. [21] proposed a filter-based method that integrates fuzzy rough sets with information theory. Their approach dynamically optimizes the weights of multiple criteria—including feature-label relevance, feature redundancy, label cooperation, and label-specific gain—thereby providing a more comprehensive evaluation of feature importance.

### B. Wrapper methods

Wrapper methods evaluate the performance of feature subsets through iterative interaction with multi-label classification (MLC) models, often achieving higher accuracy compared to filter methods. However, due to the NP-hard nature of exhaustively searching all possible feature combinations, these methods are typically computationally intensive. To overcome this challenge, recent studies have focused on incorporating heuristic and evolutionary algorithms to efficiently explore the search space and avoid local optima.

For instance, Zhang et al. [22] proposed a wrapper-based multi-label feature selection method that utilizes an improved multi-objective Particle Swarm Optimization (PSO) algorithm to identify the Pareto front of non-dominated feature subsets. Similarly, Panari et al. [23] introduced an ACO-based method that incorporates heuristic learning via reinforcement learning, modeling the search process as a Markov Decision Process (MDP) to enhance exploration efficiency and boost classification performance.

### C. Embedded methods

Embedded multi-label feature selection approaches typically incorporate regularization terms into regression models. Commonly employed regularization norms include the $\ell_1$ norm and the $\ell_{2,1}$ norm [24]. In recent years, numerous embedded feature selection methods have been developed. Huang et al. [25] proposed LLSF, which assumes that each label is associated with a specific subset of features, and that strongly correlated labels tend to share more features. Hu et al. [15] introduced SCMFS, which uses Coupled Matrix Factorization (CMF) to extract common structures between feature and label matrices while enhancing interpretability through Non-negative Matrix Factorization (NMF). Li et al. [16] proposed SLMDS, which preserves global and local label correlations using graph regularization, $\ell_{2,1}$-norm, and inner product penalties. Furthermore, Li et al. [26] developed a robust and flexible sparse regularization (RFSR) framework, leading to a global optimization method RFSFS. Fan et al. [17] presented LCIFS, which jointly exploits label correlations and feature redundancy, and later proposed LSMFS [27], which integrates a logical label matrix with a relaxed non-negative matrix to learn pseudo-labels and capture shared information among related labels, thereby mitigating the impact of sparse labels.

## III. NOTATIONS

In this section, we provide a unified explanation of the notations used throughout this paper. Bold uppercase letters denote matrices, while bold lowercase letters represent vectors. Specifically, the feature matrix is denoted as $X \in \mathbb{R}^{n \times F}$, where n is the number of samples and F is the number of features. The label matrix is denoted as $Y \in \mathbb{R}^{n \times L}$, where L represents the number of labels.

For any matrix G, we define:

- $G_{ij}$ as the element in the i-th row and j-th column of G,

- $\text{Tr}(G)$ as the trace of G (i.e., the sum of diagonal elements),

- $G^T$ as the transpose of G,
- $G^{-1}$ as the inverse of G,
- $g_i$ as the i-th row vector of G, and $\|g_i\|_2$ as the $\ell_2$ norm of $g_i$
- $\|G\|_F^2$ as the Frobenius norm of G,
- $\|G\|_{2,p}^p$ as the $\ell_{2,p}$ norm of G.

Additionally, we use $l_i$ to denote the i-th label, where $i \in 1,2 \dots L$, and $f_j$ to denote the j-th feature, where $j \in 1,2 \dots F$.

## IV. THE PROPOSED GPMFS METHOD

In this section, we propose a novel hybrid multi-label feature selection framework. This method identifies a globally optimal feature set by leveraging shared information among all labels. Building upon this, we then identify a unique personalized feature set for each label, where the features in this personalized set are specifically effective for the corresponding label. And the overall framework is depicted in Fig. 1.

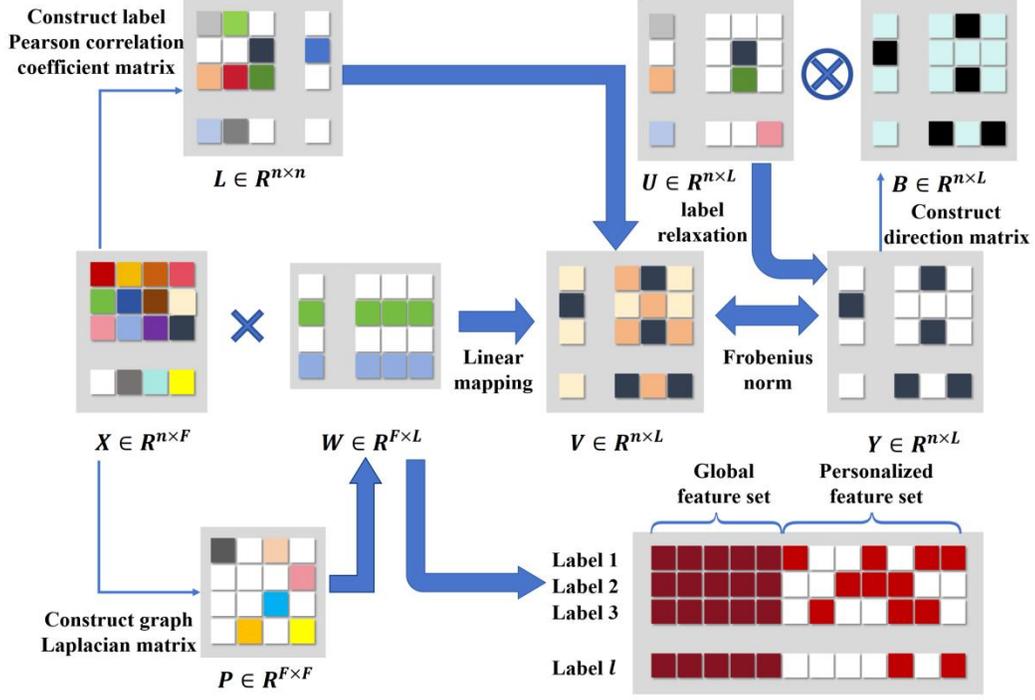

**Fig. 1.** The overall framework of GPMFS

### A. Manifold-based regression framework

Least squares regression [28] serves as a fundamental fitting method that establishes data relationships by minimizing the squared errors between predicted values and ground truth labels, finding extensive applications in multi-label feature selection. However, the conventional least squares approach exhibits significant limitations when directly applied to multi-label feature selection: its single weight matrix framework inherently restricts the model's capacity to capture complex inter-label dependencies, failing to effectively characterize label co-occurrence patterns or mutual exclusion relationships. To address these issues, this work employs an improved regression model based on manifold learning. By introducing manifold regularization terms that fully exploit the intrinsic geometric structure of the data, the model avoids degenerate solutions while explicitly modeling label correlations. This approach maintains the computational efficiency advantages of traditional least squares regression while significantly enhancing feature selection performance through improved discriminative capability and robustness. These characteristics make it particularly suitable for complex multi-label learning problems with strong label dependencies and high-dimensional feature spaces. The manifold-based regression model is formulated as follows.

$$f(X, Y, W, V) = min_{W,V} \|XW - V\|_F^2 + \|V - Y\|_F^2, \quad (1)$$

where W is the feature coefficient matrix, and the i-th row of W represents the importance of the i-th feature in approximating V.

### B. The relaxation mechanism for the label matrix Y

In multi-label learning problems, conventional binary label matrices employ rigid 0/1 encoding schemes. However, this oversimplified representation fails to accurately capture the varying degrees of association between instances and labels, nor can it effectively explore the complex correlations among labels. Moreover, the strict binary labeling tends to make models sensitive to noise. These limitations often lead to suboptimal feature selection and ultimately degrade model performance. To address these issues, we incorporate the label relaxation mechanism introduced in [27]:

$$min_{V,U} \|V - (Y + B \otimes U)\|_F^2, \quad (2)$$

where the direction matrix $B = [B_{ij}] \in \mathbb{R}^{n \times L}$ is defined as:

$$B_{ij} = \begin{cases} +1, & \text{if } Y_{ij} = 1 \\ -1, & otherwise \end{cases}, \quad (3)$$

The sign of $B_{ij}$ determines the relaxation direction: $B_{ij} = +$

1 indicates positive relaxation while $B_{ij} = -1$ denotes negative relaxation. The non-negative label relaxation matrix $U \in \mathbb{R}^{n \times L}$ controls the degree of relaxation, where $\otimes$ represents the Hadamard (element-wise) product between matrices $B$ and $U$.

### C. Structured label correlation mining via graph representation

In the process of utilizing the pseudo-label matrix $V$ to guide feature selection, we need to ensure the consistency of the local geometric structure between the multi-label data space $X$ and the pseudo-label space $V$. Specifically, if two instances $x_i$ and $x_j$ are sufficiently close in the multi-label data space $X$, then their corresponding label vectors $v_i$ and $v_j$ in the pseudo-label space $V$ should also be sufficiently close. The pseudo-label matrix $V$ is minimized as follows:

$$
\begin{aligned}
&\frac{1}{2} \Sigma_{i=1}^{n} \Sigma_{j=1}^{n} S_{ij} (v_i - v_j)^2 \\
=\ & \mathrm{Tr}(V^T(A - S)V) \\
=\ & \mathrm{Tr}(V^T L V),
\end{aligned}
\tag{4}
$$

Where $L = A - S$ represents the graph Laplacian matrix, where $A$ is a diagonal matrix with each diagonal element $A_{ii} = \Sigma_{i=1}^{n} S_{ij}$, and $S_{ij}$ denotes the pairwise similarity measure between instances $x_i$ and $x_j$ in the multi-label data space $X$. In this paper, we adopt the approach of [29] to construct a nearest neighbor graph, which effectively captures the local geometric structure in the input space $X$. The corresponding affinity graph is defined as:

$$
S_{ij} = \begin{cases} exp\left(-\frac{\|x_i - x_j\|_2^2}{\sigma^2}\right), & x_i \in \mathcal{N}_k(x_j) \text{ or } x_j \in \mathcal{N}_k(x_i), \\ 0, & otherwise \end{cases}
\tag{5}
$$

where $\mathcal{N}_k(x_i)$ is the set of $k$-nearest neighbors for instance $x_i$.

### D. Feature redundancy reduction

When two features are highly correlated, selecting both may lead to redundancy and unnecessary computational costs. To mitigate this issue, we introduce a feature redundancy regularization term inspired by [17]:

$$
r(W) = \frac{1}{F(F-1)} \Sigma_{i=1}^{F} \|w_i\|_2 \Sigma_{j=1, j \neq i}^{F} \|w_j\|_2 P_{ij},
\tag{6}
$$

where $P_{ij}$ represents the Pearson correlation coefficient between the ith and jth features, and $\frac{1}{F(F-1)}$ is a normalization factor.

Furthermore, to eliminate ineffective redundant features, we impose a sparsity regularization on the feature selection matrix $W$:

$$
\Omega(W) = \Sigma_{i=1}^{d} \|w_i\|_2^p = \|W\|_{2,p}^p.
\tag{7}
$$

### E. The objective function

Considering the optimization mechanisms discussed in sections 3.1 to 3.4, the objective function of the multi-label selection method we propose can be described as follows:

$$
\begin{aligned}
\min_{W,V,U} \ & \|XW - V\|_F^2 \\
& + \alpha \|V - (Y + B \otimes U)\|_F^2 + \beta T_r V^T L V \\
& + \gamma \Sigma_{i,j=1}^{\alpha} \|w_i\|_2 \|w_j\|_2 P_{ij} + \eta \|W\|_{2,p}^p.
\end{aligned}
\tag{8}
$$

The model incorporates four non-negative parameters: $\alpha, \beta, \gamma,$ and $\lambda$. Using the optimization method described in the next chapter, we obtain the optimal feature selection matrix $W$. Subsequently, we sort the $\|w_i\|_2$ norms (where $i = 1, ..., F$) in descending order and select the top-ranked features to construct the global feature sets GF.

### E. Select personalized features for each label

Existing multi-label feature selection methods typically optimize a unified objective function to select a global optimal feature set for all labels. These approaches focus solely on the $L_2$-norm $\|w_i\|_2$ as an indicator of each feature's average importance across all labels, ultimately yielding the global feature set GF described in Section 3.5. However, such methods inherently neglect label-specific characteristics. Specifically, when a feature fj is critically important for predicting a particular label li but irrelevant to other labels, it will likely be excluded from the global feature set GF. To analyze this phenomenon mathematically, we examine the expression of $\|w_j\|_2$:

$$
\|w_j\|_2 = \sqrt{w_{j1}^2 + w_{j2}^2 + ... + w_{ji}^2 + ... + w_{jL}^2},
\tag{9}
$$

here, $|w_{ji}|$ represents the importance of the j-th feature for predicting the i-th label. For a feature fj that is highly relevant to only one label li, its corresponding $|w_{ji}|$ may be significant, while all other $|w_{jk}|$ (where $k \neq i$) values remain negligible. Consequently, the overall $\ell_2$-norm $\|w_j\|_2$ of such features tends to be small, causing them to be omitted from GF.

To address this limitation, we propose a novel paradigm that selects personalized features for each label $l_i$. Specifically, we aim to identify feature $f_j$ with large $|w_{ji}|$ values for a specific label $l_i$ but modest $\|w_j\|_2$ norms globally. For features $f_j$ not selected in GF, we include it in the personalized feature $PF_i$ for label $l_i$ if $f_j$ satisfies:

$$
\Sigma_{l_r \in GF} \mathbb{I}\left(|w_{ij}| > \frac{\|w_r\|_2}{\sqrt{L}}\right) > q \cdot |GF|,
\tag{10}
$$

here, $\mathbb{I}(\cdot)$ denotes the indicator function, which returns 1 if the condition inside holds, and 0 otherwise. $\sqrt{L}$ serves as a normalization factor to account for the scale of $\|w_k\|_2$, while q acts as a proportional threshold to control the selectivity of personalized features.

The complete feature set $CF_i$ for each label $l_i$ is then constructed by combining the global feature set GF and the personalized feature set $PF_i$.

## V. OPTIMIZATION SOLUTION

In this section, we present the optimization approach for our objective function. For notational convenience, we denote our objective function as $\Phi$, which is defined as:

$$
\begin{aligned}
\Phi(W,V,U) = \ & \|XW - V\|_F^2 \\
& + \alpha \|V - (Y + B \otimes U)\|_F^2 + \beta T_r V^T L V \\
& + \gamma \Sigma_{i,j=1}^{\alpha} \|w_i\|_2 \|w_j\|_2 P_{ij} + \lambda \|W\|_{2,p}^p.
\end{aligned}
\tag{11}
$$

The feature correlation term $\Sigma_{i,j=1}^{\alpha} \|w_i\|_2 \|w_j\|_2 P_{ij}$ is non-convex, and the sparsity regularization term $\|W\|_{2,p}^p$ becomes non-convex when $0 < p < 1$. To address these challenges, we employ an alternating optimization algorithm. This approach sequentially optimizes each variable while keeping the others fixed. Specifically, we alternately update the variables $W, V,$ and $U$. In the following subsections, we will detail the optimization procedures for each of these variables.

### A. Update W as fixed V and U

When V and U are held fixed, we simplify the objective function $\Phi$ by removing terms independent of W, resulting in:

$$\Theta = \|XW - V\|_F^2 + \gamma \sum_{i,j=1}^{\alpha} \|w_i\|_2 \|w_j\|_2 P_{ij} + \eta \|W\|_{2,p}^p. \quad (12)$$

Taking the derivative of $\Theta$ with respect to W, we get

$$\frac{\partial \Theta}{\partial W} = 2[X^T(XW - V) + \gamma MW + \eta NW]. \quad (13)$$

The matrices M and N are both diagonal matrices with their respective diagonal elements defined as:

$$M_{ii} = \frac{\sum_j \|w_j\|_2 P_{ij}}{2\|w_i\|_2}, \quad (14)$$

$$N_{ii} = \frac{p}{2\|w_i\|_2^{2-p}}. \quad (15)$$

By setting the derivative to zero, we obtain the closed-form solution for W:

$$W = (X^TX + \gamma M + \eta N)^{-1} X^TV. \quad (16)$$

### B. Update V as fixed W and U

When W and U are held fixed, we simplify the objective function $\Phi$ by removing terms independent of V, resulting in:

$$\Theta = \|XW - V\|_F^2 + \alpha\|V - (Y + B \otimes U)\|_F^2 + \beta T_r V^T LV. \quad (17)$$

Taking the derivative of $\Theta$ with respect to V, we get

$$\frac{\partial \Theta}{\partial V} = -2(XW - V) + 2\alpha(V - (Y + B \otimes U)) + \beta(LV + L^TV), \quad (18)$$

given that L is a symmetric matrix ($L = L^T$), we compute the partial derivative of $\Theta$ with respect to V :

$$\frac{\partial \Theta}{\partial V} = 2[-(XW - V) + \alpha(V - (Y + B \otimes U)) + \beta LV]. \quad (19)$$

By setting the derivative to zero, we obtain the closed-form solution for V:

$$V = ((1 + \alpha)I + \beta L)^{-1}(XW + \alpha(Y + B \otimes U). \quad (20)$$

### C. Update U as fixed W and V

When W and V are held fixed, we simplify the objective function $\Phi$ by removing terms independent of , resulting in:

$$\Theta = \alpha \| V - (Y + B \otimes U)\|_F^2 \quad (21)$$

The optimal solution for U can be obtained through an element-wise thresholding operation, as established in [30]:

$$U = max(B \otimes (V - Y), 0). \quad (22)$$

Having derived the optimization procedures for W, V, and U respectively, we now present the pseudocode of our proposed GPMFS algorithm.

---

**Algorithm 1:** The proposed method **GPMLS**

**Input:** The feature matrix $X \in \mathbb{R}^{n \times F}$ , the label matrix $Y \in \mathbb{R}^{n \times L}$ , and parameters $\alpha, \beta, \gamma, \lambda, p$ and $q$.

**Output:** For each label $l_i$, select the features from its global feature set ($GF$) and its personalized feature set ($PF_i$).

1   Set the iteration step $t = 0$, and initialize $W^t \in \mathbb{R}^{F \times L}, V^t \in \mathbb{R}^{n \times L}$, and $U^t \in \mathbb{R}^{n \times L}$;

2   Construct the Pearsons correlation matrix $P$ and the graph laplacian matrix $L$;

3   while not converged do

4     Calculate the matrix $M^t \leftarrow diag(M_{11}^t, ..., M_{dd}^t)$ by (14);

5     Calculate the matrix $N^t \leftarrow diag(N_{11}^t, ..., N_{dd}^t)$ by(15);

6     $A^t = X^TX + \gamma M^t + \lambda N^t$;

7     $W^{t+1} = (A^t)^{-1}X^TV$;

8     $D = Y + B \otimes U^t$;

9     $V^{t+1} = ((1 + \alpha)I + \beta L)^{-1}(XW^{t+1} + \alpha D)$;

10    $U^{t+1} = max(B \otimes (V^{t+1} - Y), 0)$;

11    $t = t + 1$;

12   end

13   Sort all features according to the descending order of $\|w_i\|_2(i = 1, ..., F)$ and select top $k$ percent features as global feature sets $GF$.

14   For each label $l_i$, $i \in 1, 2 ... L$:

15     For feature $f_j$ not in set $GF$, $j \in 1, 2 ... F$:

16      if $\sum_{l_i \in GF} \mathbb{I}\left(|w_{ij}| > \frac{\|w_i, l_i\|}{\sqrt{L}}\right) > q \cdot |GF|$:

17       select feature $f_j$ as a personolized feture for label $l_i$

18   Obtain the personalized feature set $PF_i$ for each label $l_i$

---

+-

## VI. Experiments and results

In this section, we conduct a series of comprehensive experiments to systematically evaluate the performance of the proposed algorithm. Specifically, we benchmark the proposed method against state-of-the-art multi-label feature selection methods on multiple datasets, evaluating performance using Hamming Loss, Micro-F1, One-error, Average Precision and Macro-F1. Furthermore, we investigate the impact of the personalized feature selection threshold q on multi-label prediction performance. Additionally, we perform parameter sensitivity analysis to validate the robustness of our model. Finally, we provide a detailed discussion on the convergence and computational complexity of the proposed method, supported by both theoretical analysis and empirical validation.

### A. Datasets

In our experiments, we utilized 10 publicly available multi-label datasets to evaluate the performance of the proposed method, all of which can be accessed from the Mulan Library. These datasets cover various domains, including music (Emotions), biology (Yeast), images (Flags and Scene), and text (Business, Computers, Education, Enron, Medical, and Social). The detailed descriptions of these datasets are presented in TABLE I, where "card" denotes the average number of labels per sample, and "Den" represents label density.

TABLE I
THE DETAILS OF DATASETS

| Datasets | #Instances | #Features | #Labels | Card | Den | Domain |
|----------|-----------|-----------|---------|------|-----|--------|
| Business | 5000 | 438 | 30 | 1.588 | 0.053 | Text |
| Computers | 5000 | 681 | 33 | 1.508 | 0.046 | Text |
| Education | 5000 | 550 | 33 | 1.461 | 0.044 | Text |
| Emotions | 593 | 72 | 6 | 1.869 | 0.311 | Music |
| Enron | 1702 | 1001 | 53 | 3.378 | 0.064 | Text |
| Flags | 194 | 19 | 7 | 3.392 | 0.485 | Image |
| Medical | 978 | 1449 | 45 | 1.245 | 0.028 | Text |
| Scene | 2407 | 294 | 6 | 1.074 | 0.179 | Image |
| Social | 5000 | 1047 | 39 | 1.283 | 0.033 | Text |
| Yeast | 2417 | 103 | 14 | 4.237 | 0.303 | Biology |

*B. Evaluation Metrics*

In this experiment, we employ multiple evaluation metrics specifically designed for multi-label classification to comprehensively assess the performance of the proposed method. These metrics include Hamming Loss, Micro-F1, One-error, Average Precision, and Macro-F1, each evaluating different aspects of the model's predictive ability and label assignment quality.

- Hamming Loss (HL): Measures the proportion of incorrectly predicted labels. Lower values indicate better performance.

- Micro-F1: Balances precision and recall by aggregating TP, FP, and FN across all labels, making it suitable for imbalanced datasets.

- One-error: Checks if the top-ranked predicted label is in the true label set. Lower values indicate better ranking accuracy.

- Average Precision (AP): Computes the mean precision over relevant labels, assessing how well correct labels are ranked higher.

- Macro-F1: Calculates F1 scores for each label separately and averages them, useful for imbalanced multi-label tasks. During the calculation of Macro-F1, if a label is absent in either the ground truth or the predictions (i.e., resulting in a zero denominator), its F1 score is automatically set to 1 by default.

The computational methods for these evaluation metrics are described in detail in [31]. By using these metrics, we can perform a comprehensive quantitative analysis of the model's performance, ensuring the reliability and robustness of the experimental results.

*C. Compared methods*

In this study, we compare our proposed method GPMFS with six state-of-the-art multi-label feature selection approaches, namely FIMF, SCMFS, RFSFS, LSMFS, SLMDS, and LCIFS. A brief introduction to each of these methods is provided below.

- GPMFS: The regularization parameters $\alpha, \beta, \gamma$ and $\lambda$ cover the interval from $\{10^{-2}, 10^{-1}, 1, 10^1, 10^2\}$, and the norm parameter p and the threshold parameter q are both set to 0.8.

- FIMF [18]: a computationally efficient method that ranks features based on information-theoretic measures to identify an optimal subset for improving multi-label learning accuracy.

- SCMFS [15]: a feature selection method that leverages Coupled Matrix Factorization (CMF) to capture shared structures between feature and label matrices while enhancing interpretability with Non-negative Matrix Factorization (NMF). The regularization parameters $\alpha, \beta,$ and $\gamma$ take values from $\{10^{-2}, 10^{-1}, 1, 10^1, 10^2\}$.

- RFSFS [26]: a feature selection method that employs Robust Flexible Sparse Regularization (RFSR) and an alternating multipliers-based optimization scheme to enhance sparsity, discrimination, and redundancy reduction in multi-label learning. The regularization parameters $\alpha, \beta,$ and $\gamma$ are span the range from $\{10^{-2}, 10^{-1}, 1, 10^1, 10^2\}$.

- LSMFS [27]: a feature selection method that constructs a pseudo-label matrix via label relaxation and captures shared label information to mitigate low-density label influence in multi-label learning. The regularization parameters $\alpha, \beta, \gamma$ and $\lambda$ cover the interval from $\{10^{-2}, 10^{-1}, 1, 10^1, 10^2\}$.

- SLMDS [16]: a feature selection method that preserves global and dynamic local label correlations while enforcing sparsity and reducing redundancy. The regularization parameters $\alpha, \beta,$ and $\gamma$ are varied from $\{10^{-2}, 10^{-1}, 1, 10^1, 10^2\}$.

- LCIFS [17]: a feature selection method that incorporates label correlations and controls feature redundancy using a manifold regression model and $\ell_{2,p}$ norm regularization. The regularization parameters $\alpha, \beta, \gamma$ and $\lambda$ are set within the range from $\{10^{-2}, 10^{-1}, 1, 10^1, 10^2\}$, and the parameter of norm p = 0.8.

In our comparative experiments, we employ MLKNN (k=10) [32] for multi-label prediction. As the optimal feature set for each label is personalized in our proposed method, we instead leverage KNN (k=10) to independently predict each label. Notably, MLKNN also independently computes the posterior probability for each label during the prediction process, inherently treating each label separately. Therefore, under the same parameter settings, both methods share a consistent prediction mechanism, ensuring the comparability of the experimental results. To approximate the optimal solution while maintaining computational efficiency, we adopt a two-stage neighbor selection strategy for per-label KNN prediction. Specifically, we first identify the top 20 nearest neighbors using a globally shared feature set across all labels. Then, for each label, we select the final 10 neighbors from this candidate pool based on the corresponding label-specific personalized feature subset. This approximation effectively balances predictive performance and computational cost, enabling efficient individualized label prediction under the framework of personalized feature selection. Additionally, we employ five-fold cross-validation in our experiments, and all reported results are presented as average values across the folds.

*D. Results and analysis*

To evaluate the performance of all compared methods, we redefine the feature space by selecting the top 20% of features based on their weights, as suggested in [17][27]. And TABLE II-VI present the optimal performance achieved by each method, with the corresponding parameters selected through the grid search strategy [33]. The optimal results for all metrics are emphasized by making them bold and underlined in the table. Additionally, the Win/Tie/Loss records represent the number of datasets where GPMFS outperforms, equals, or performs worse than the other compared methods. It can be observed that GPMFS performs best in the majority of cases. The following provides a detailed analysis of the experimental results.

In terms Hamming Loss, GPMFS achieves the best performance on 10 datasets. Specifically, it shows the largest reduction of 5% in Hamming Loss on the Flags dataset. On the Enron dataset, GPMFS's performance is comparable to LSMFS. For Micro-F1, GPMFS achieves the best result on 8 datasets. The improvements on Flags, Emotions, and Scene are 5%, 5%, and 4%, respectively. However, it performs slightly worse than RFSFS on the Education dataset and slightly worse than LCIFS on the Social dataset. Regarding One-error, GPMFS performs the best across all datasets, with the most notable reduction on the Education dataset, where One-error decreases by 23%. For Average Precision, GPMFS achieves the best result on 10 datasets. It shows a 9% improvement on Education and a 5% improvement on

TABLE II
RESULTS OF OF DIFFERENT FEATURE SELECTION ALGORITHMS IN TERMS OF HL

| Data set | Hamming Loss ↓ | | | | | | |
|---|---|---|---|---|---|---|---|
| | FIMF | SCMFS | RFSFS | LSMFS | SLMDS | LCIFS | GPMFS |
| Business | 0.0289 | 0.0265 | 0.0265 | 0.0263 | 0.0265 | 0.0263 | **<u>0.0242</u>** |
| Computers | 0.0394 | 0.0385 | 0.0385 | 0.0368 | 0.0377 | 0.0367 | **<u>0.0339</u>** |
| Education | 0.0434 | 0.0399 | 0.0399 | 0.0398 | 0.0398 | 0.0395 | **<u>0.0372</u>** |
| Emotions | 0.2406 | 0.2095 | 0.2052 | 0.1995 | 0.2032 | 0.1993 | **<u>0.1737</u>** |
| Enron | 0.0520 | 0.0513 | 0.0511 | **<u>0.0506</u>** | 0.0507 | 0.0509 | **<u>0.0506</u>** |
| Flags | 0.2923 | 0.2644 | 0.2511 | 0.2511 | 0.2511 | 0.2371 | **<u>0.1905</u>** |
| Medical | 0.0153 | 0.0146 | 0.0142 | 0.0116 | 0.0138 | 0.0141 | **<u>0.0086</u>** |
| Scene | 0.1525 | 0.1018 | 0.0989 | 0.0897 | 0.0971 | 0.0916 | **<u>0.0764</u>** |
| Social | 0.0259 | 0.0225 | 0.0224 | 0.0229 | 0.0238 | 0.0235 | **<u>0.0207</u>** |
| Yeast | 0.2232 | 0.2084 | 0.2056 | 0.2091 | 0.2117 | 0.2022 | **<u>0.1936</u>** |
| Win/Tie/Loss | 10/0/0 | 10/0/0 | 10/0/0 | 9/1/0 | 10/0/0 | 10/0/0 | |

TABLE III
RESULTS OF OF DIFFERENT FEATURE SELECTION ALGORITHMS IN TERMS OF MICRO-F1

| Data set | Micro-F1 ↑ | | | | | | |
|---|---|---|---|---|---|---|---|
| | FIMF | SCMFS | RFSFS | LSMFS | SLMDS | LCIFS | GPMFS |
| Business | 0.6792 | 0.7113 | 0.7113 | 0.7159 | 0.7122 | 0.6967 | **<u>0.7379</u>** |
| Computers | 0.4179 | 0.4810 | 0.4818 | 0.4831 | 0.4821 | 0.4847 | **<u>0.5051</u>** |
| Education | 0.2466 | 0.3657 | **<u>0.3883</u>** | 0.3736 | 0.3661 | 0.3677 | 0.3731 |
| Emotions | 0.5800 | 0.6502 | 0.6515 | 0.6583 | 0.6552 | 0.6651 | **<u>0.7123</u>** |
| Enron | 0.5114 | 0.5130 | 0.5138 | 0.5181 | 0.5144 | 0.5153 | **<u>0.5196</u>** |
| Flags | 0.6914 | 0.7250 | 0.7437 | 0.7437 | 0.7437 | 0.7547 | **<u>0.8060</u>** |
| Medical | 0.6908 | 0.7648 | 0.7976 | 0.7872 | 0.7507 | 0.7792 | **<u>0.8369</u>** |
| Scene | 0.5009 | 0.6974 | 0.7001 | 0.7345 | 0.7008 | 0.7037 | **<u>0.7788</u>** |
| Social | 0.4870 | 0.5853 | 0.5859 | 0.5883 | 0.5861 | **<u>0.5984</u>** | 0.5889 |
| Yeast | 0.5926 | 0.6236 | 0.6235 | 0.6233 | 0.6239 | 0.6388 | **<u>0.6673</u>** |
| Win/Tie/Loss | 10/0/0 | 10/0/0 | 9/0/1 | 9/0/1 | 10/0/0 | 9/0/1 | |

TABLE IV
RESULTS OF OF DIFFERENT FEATURE SELECTION ALGORITHMS IN TERMS OF ONE-ERROR

| Data set | One-error ↓ | | | | | | |
|---|---|---|---|---|---|---|---|
| | FIMF | SCMFS | RFSFS | LSMFS | SLMDS | LCIFS | GPMFS |

| | | | | | | |
|---|---|---|---|---|---|---|
| Business | 0.1656 | 0.1582 | 0.1575 | 0.1562 | 0.1573 | 0.1628 | **_0.0099_** |
| Computers | 0.4980 | 0.4762 | 0.4739 | 0.4741 | 0.4728 | 0.4476 | **_0.3640_** |
| Education | 0.7568 | 0.7328 | 0.7294 | 0.7288 | 0.7286 | 0.7336 | **_0.4952_** |
| Emotions | 0.4536 | 0.3491 | 0.3676 | 0.3424 | 0.3558 | 0.2773 | **_0.2101_** |
| Enron | 0.3713 | 0.3173 | 0.3196 | 0.3032 | 0.3555 | 0.3252 | **_0.2812_** |
| Flags | 0.2169 | 0.2008 | 0.1960 | 0.1969 | 0.2006 | 0.0789 | **_0.0782_** |
| Medical | 0.3701 | 0.3261 | 0.2285 | 0.2352 | 0.2249 | 0.2416 | **_0.1128_** |
| Scene | 0.6967 | 0.3897 | 0.3868 | 0.3299 | 0.3612 | 0.2967 | **_0.1784_** |
| Social | 0.5134 | 0.4520 | 0.4503 | 0.4424 | 0.4498 | 0.4142 | **_0.3093_** |
| Yeast | 0.2751 | 0.2644 | 0.2625 | 0.2645 | 0.2582 | 0.2397 | **_0.2052_** |
| Win/Tie/Loss | 10/0/0 | 10/0/0 | 10/0/0 | 10/0/0 | 10/0/0 | 9/1/0 | |

TABLE V
RESULTS OF OF DIFFERENT FEATURE SELECTION ALGORITHMS IN TERMS OF AP

| Data set | Average Precision ↑ | | | | | | |
|---|---|---|---|---|---|---|---|
| | FIMF | SCMFS | RFSFS | LSMFS | SLMDS | LCIFS | GPMFS |
| Business | 0.8331 | 0.8297 | 0.8313 | 0.8429 | 0.8326 | 0.8360 | **_0.8737_** |
| Computers | 0.5824 | 0.5962 | 0.5964 | 0.6019 | 0.5969 | 0.6046 | **_0.6416_** |
| Education | 0.4180 | 0.4284 | 0.4347 | 0.4422 | 0.4328 | 0.4427 | **_0.5301_** |
| Emotions | 0.6831 | 0.7418 | 0.7402 | 0.7478 | 0.7445 | 0.7831 | **_0.8071_** |
| Enron | 0.5613 | 0.6111 | 0.6149 | 0.6224 | 0.6167 | 0.6219 | **_0.6421_** |
| Flags | 0.7940 | 0.8021 | 0.8060 | 0.8074 | 0.8036 | 0.8570 | **_0.8645_** |
| Medical | 0.7233 | 0.7490 | 0.8315 | 0.8056 | 0.8106 | 0.8092 | **_0.8826_** |
| Scene | 0.5657 | 0.7746 | 0.7735 | 0.8055 | 0.7901 | 0.8227 | **_0.8669_** |
| Social | 0.6143 | 0.6497 | 0.6528 | 0.6741 | 0.6533 | 0.6693 | **_0.7073_** |
| Yeast | 0.6791 | 0.6875 | 0.6857 | 0.6949 | 0.6923 | 0.7113 | **_0.7483_** |
| Win/Tie/Loss | 10/0/0 | 10/0/0 | 10/0/0 | 9/0/1 | 10/0/0 | 10/0/0 | |

TABLE VI
RESULTS OF OF DIFFERENT FEATURE SELECTION ALGORITHMS IN TERMS OF MACRO-F1

| Data set | Macro-F1 ↑ | | | | | | |
|---|---|---|---|---|---|---|---|
| | FIMF | SCMFS | RFSFS | LSMFS | SLMDS | LCIFS | GPMFS |
| Business | 0.2709 | 0.3338 | 0.3335 | 0.3547 | 0.3386 | 0.3201 | **_0.4081_** |
| Computers | 0.2462 | 0.3392 | 0.3393 | 0.3349 | 0.3396 | 0.3437 | **_0.4256_** |
| Education | 0.2655 | 0.3648 | 0.3391 | 0.3394 | 0.3516 | 0.3344 | **_0.3928_** |
| Emotions | 0.5508 | 0.6279 | 0.6274 | 0.6376 | 0.6283 | 0.6745 | **_0.6941_** |
| Enron | 0.2772 | 0.2966 | 0.3013 | 0.3040 | 0.3016 | 0.3049 | **_0.3112_** |
| Flags | 0.6394 | 0.6452 | 0.6505 | 0.6452 | 0.6505 | 0.6616 | **_0.8419_** |
| Medical | 0.5845 | 0.6189 | **_0.6706_** | 0.6476 | 0.6309 | 0.6344 | 0.6599 |
| Scene | 0.5032 | 0.7005 | 0.7091 | 0.7409 | 0.7124 | 0.7511 | **_0.7712_** |
| Social | 0.2942 | 0.3276 | 0.3476 | **_0.3501_** | 0.3296 | 0.3280 | 0.3374 |
| Yeast | 0.4005 | 0.4231 | 0.4275 | 0.4298 | 0.4245 | 0.5551 | **_0.5726_** |
| Win/Tie/Loss | 10/0/0 | 10/0/0 | 8/0/2 | 9/0/1 | 10/0/0 | 10/0/0 | |

Medical. Finally, for Macro-F1, GPMFS performs best on 8 datasets. On Medical and Social, its performance lags behind RFSFS and LSMFS, while it shows the most significant improvements on Flags and Computers, with increases of 18% and 8%, respectively.

GPMFS demonstrates outstanding performance on most multi-label datasets, particularly on benchmarks such as Business, Enron, and Yeast, where it shows significant improvements in predictive accuracy. This improvement can be attributed to the proposed global-personalized feature selection mechanism, which effectively integrates the shared structure across labels with the discriminative power of label-specific features. However, its performance on the Social dataset is relatively modest, and in some cases even inferior to certain baseline methods. This observation highlights the limitations of GPMFS when confronted with specific data characteristics.

From a dataset perspective, Social has 39 labels—placing it in the upper-middle range in terms of label count—but exhibits an extremely low label density (Den = 0.033), indicating that labels are highly sparse, with each instance associated with only about 1.283 labels on average (Card). In such scenarios, GPMFS may struggle to learn effective personalized features due to insufficient training samples for each label, which can result in the selection of non-discriminative or noisy features, ultimately impairing model performance.

In contrast, the Emotions dataset, although it contains fewer labels (6), has a much higher label density (0.311), with each instance associated with nearly two labels on average. This higher co-occurrence facilitates stronger inter-label relationships, allowing the global feature component to effectively capture shared structures. As a result, even with a smaller proportion of personalized features, GPMFS can still achieve competitive performance.

Furthermore, the Social dataset has the highest feature dimensionality among all datasets considered (1047 features), contributing to an inherently complex input space. In such high-dimensional settings, if the personalized feature selection component fails to sufficiently filter out redundant features or allocates excessive noisy features to low-frequency labels, the model may overfit to a small subset of label-specific patterns in the training set, thereby reducing generalization ability. By contrast, the Yeast dataset, despite having 14 labels and a high average cardinality (4.237), contains only 103 features. In this lower-dimensional space, GPMFS is more capable of extracting discriminative personalized features, leading to a more pronounced performance gain.

In summary, while GPMFS excels in a wide range of multi-label learning tasks by jointly modeling label correlations and label-specific distinctions, its performance may be constrained in scenarios characterized by extreme label sparsity and high-dimensional feature spaces, as exemplified by the Social dataset.

To rigorously assess the performance disparities among multiple competing algorithms, a two-phase statistical testing framework is adopted. Table 7 presents the Friedman statistics ($F_F$) [34] and the corresponding critical values at $\alpha = 0.05$ significance level. Notably, all $F_F$ values substantially exceed the critical value of 2.2720, with particularly prominent results in Average Precision ($F_F$ = 51.5769) and Hamming Loss ($F_F$ = 36.4151). This overwhelming evidence leads to the rejection of the null hypothesis, confirming statistically significant performance differences among the methods.

TABLE VII
FRIEDMAN TEST STATISTICS AND CRITICAL VALUE ACROSS EVALUATION METRICS

| Evaluation metric | $F_F$ | Critical value |
|---|---|---|
| Hamming Loss | 36.4151 | |
| Micro-F1 | 27.6471 | |
| One-error | 23.2251 | 2.2720 |
| Average Precision | 51.5769 | |
| Macro-F 1 | 17.8556 | |

Based on these results, we proceeded with post-hoc analysis using the Bonferroni-Dunn test [34] to examine performance differences between our proposed GPMFS method and other comparative approaches. The significance of differences between the control method (GPMFS) and each comparison method was determined using the critical difference (CD) metric:

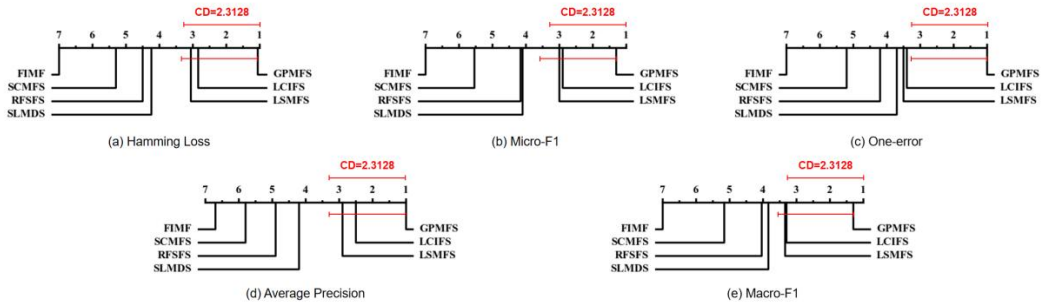

Fig. 2. Comparison of the control method (GPMFS) with six other methods using the Bonferroni-Dunn test across all evaluation metrics, with CD= 2.3128 at a significance level of α = 0.1.

$$CD = q_\alpha \sqrt{\frac{c(c+1)}{6k}}, \qquad (23)$$

where $q_\alpha = 2.394$ at the significance level $\alpha = 0.1$, c denotes the total number of compared methods, and k refers to the number of datasets used in the experiment. In our experiments, with c = 7 and k = 10, the critical difference (CD) value is determined to be 2.3128. As shown in Fig. 2, the CD diagrams depict the comparative rankings of various methods across multiple evaluation metrics, including

Hamming Loss, Micro-F1, One-error, Average Precision, and Macro-F1. The CD value of 2.3128 serves as a statistical threshold, indicating that if the ranking difference between two methods is smaller than this threshold, their performance differences are not statistically significant.

From the experimental results, the proposed method GPMFS exhibits a statistically significant difference from all other methods in One-error and demonstrates significant differences from FIMF, SCMFS, RFSFS, and SLMDS across other evaluation metrics. Moreover, GPMFS consistently achieves the best ranking across all five metrics, underscoring its superior performance. These findings highlight the significant advantages of GPMFS across all evaluation metrics, further validating its effectiveness as a robust multi-label feature selection method compared to state-of-the-art approaches.

*E. Impact of the personalized feature selection threshold q*

The threshold $q$ proportionally regulates the selection of personalized features. A larger $q$ imposes stricter conditions for selecting personalized features. In our experiments, we set $q = 0.5$.

First, for each dataset, we calculated the average proportion of personalized feature sets across all labels, which ranges from 2.1% to 7.9%, as shown in Fig. 3.

To further validate the advantage of our method, we conducted comparative experiments on the Emotions and Scene datasets. To eliminate the influence of feature quantity on the final results, we adjusted the baseline methods to select the same number of features as our approach, specifically fixing the selection at 23.3% and 25% for the two datasets, respectively.

As shown in Fig. 4, when the total number of selected features is controlled, our proposed method, GPMFS,

**Fig. 3.** Preportion of selecter features in each dataset.

achieves the best performance across all metrics on both Emotions and Scene datasets. This ensures a more accurate evaluation of whether the superiority of our method stems from selecting personalized important features for each label rather than merely increasing the total number of features.

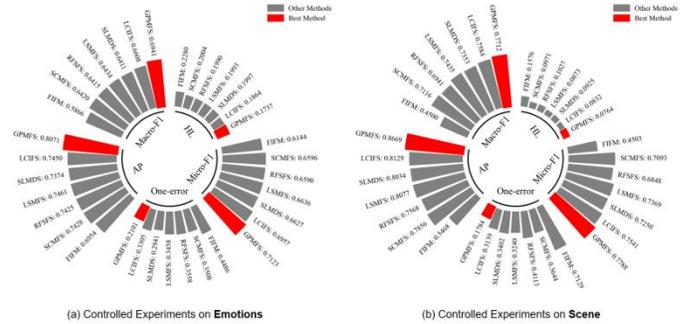

**Fig. 4.** Comparative performance of GPMFS and six other methods across all metrics on the Emotions and Scene datasets, with fixed feature selection percentages (23.3% and 25%).

Finally, we investigated how each evaluation metric varies with the personalized feature selection threshold q. As visible in Figure 5(a), when $q = 1.0$, the personalized feature set for each label is empty, whereas when $q = 0$, the number of personalized features per label reaches its maximum. As shown in Figure 5(b)~(f), we plotted the variation of evaluation metrics across five datasets as q changes. In our experiments, we set $\alpha$, $\beta$, $\gamma$, and $\lambda$ to 1.0, and all reported results represent the average over 20 repeated runs. Since the scales of the same metric vary across different datasets, we applied Min-Max scaling to standardize all metrics and better capture their trends.

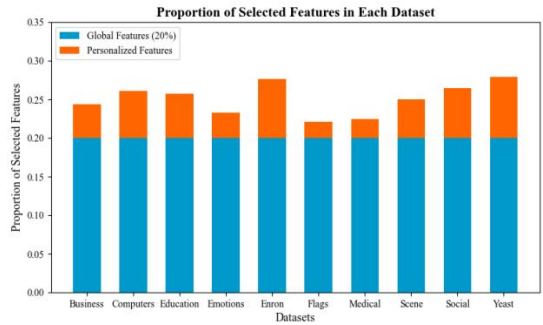

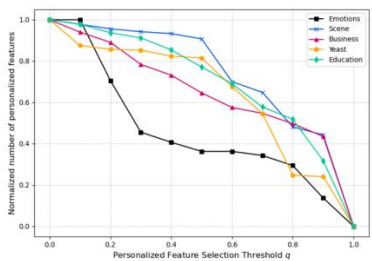

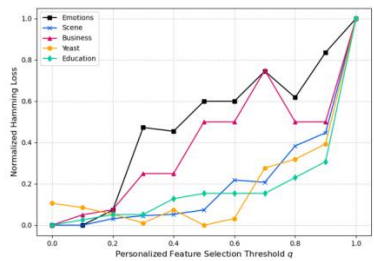

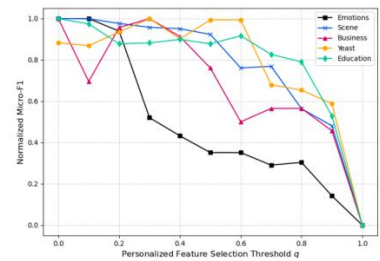

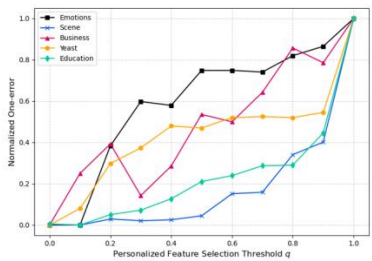

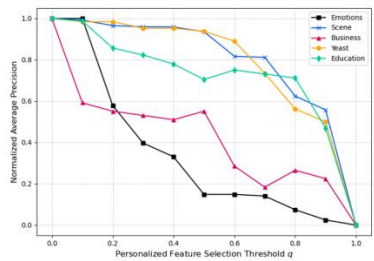

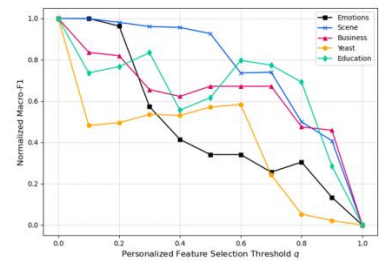

**Fig. 5.** Impact of varying the personalized threshold q on the number of personalized features and evaluation metrics.

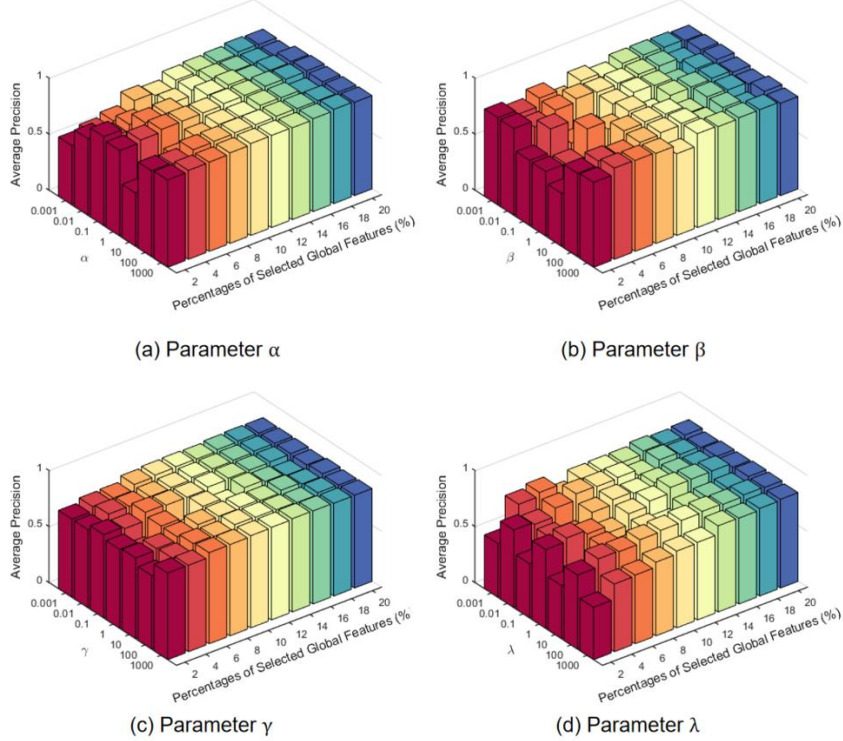

(a) Parameter α  (b) Parameter β

(c) Parameter γ  (d) Parameter λ

**Fig. 6.** Sensitivity analysis of parameters α, β, γ, and λ on AP under varying feature selection proportions on the Scene dataset

In our experiments, we observe that as q gradually increases within the range [0,1], the number of personalized features decreases, and multi-label prediction performance also declines. This trend demonstrates the significant impact of personalized features on multi-label learning, thereby highlighting the effectiveness and importance of our proposed method. While a smaller q can enhance the model's expressiveness for certain labels, it may also lead to significant feature redundancy, increased overfitting risk, and higher computational cost. In our experiments, We select q = 0.5 based on two key considerations: (1) it significantly improves predictive performance without introducing excessive redundancy, and (2) it strikes a favorable balance between global shared modeling and labelspecific customization, enhancing the model's robustness and generalization ability. In conclusion, setting q = 0.5 represents the optimal balance between predictive performance and feature selection efficiency.

*F. Parameter sensitivity analysis*

To investigate the influence of the parameters α, β, γ, and λ in the GPMFS algorithm on model performance, we conducted a parameter sensitivity analysis on the Scene dataset. Each parameter was individually varied within the search space $\{10^{-3}, 10^{-2}, 10^{-1}, 10^{0}, 10^{1}, 10^{2}, 10^{3}\}$, while keeping the remaining three parameters fixed at 1. To comprehensively evaluate the impact of these hyperparameters on Average precision (AP), we varied the proportion of selected features from 2% to 20% in increments of 2%. The experimental results are illustrated in Fig. 6.

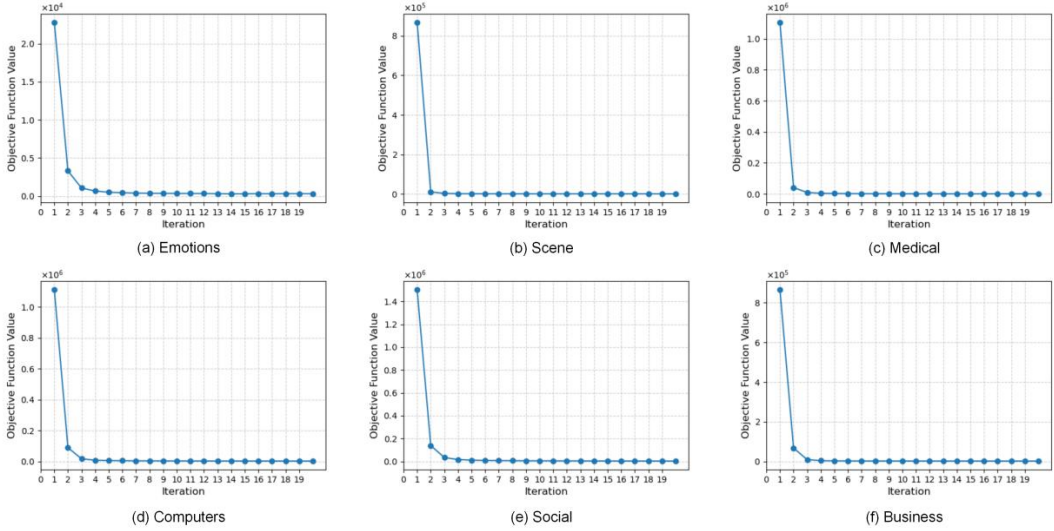

**Fig. 7.** Convergence curves of GPMFS across six benchmark datasets

The experimental results reveal that GPMFS exhibits stable and steadily improving performance as the proportion of selected features increases, underscoring its capability to capture informative features effectively. When the selected feature ratio is low, the method shows some degree of sensitivity to variations in individual hyperparameters. Nonetheless, within a broad and practical range of parameter values, the performance remains competitive. As the feature subset grows, the influence of parameter tuning becomes increasingly limited, suggesting that GPMFS is not only effective but also robust to parameter fluctuations.

*G. Convergence and time complexity analysis*

To evaluate the convergence behavior of GPMFS, we conducted experiments on six benchmark datasets(Emotions, Scene, Medical, Computers, Social and Business). In the convergence analysis, the number of iterations for GPMFS was fixed at 20. Fig. 7 presents the convergence curves for different datasets, where the x-axis and y-axis represent the number of iterations and the objective function value, respectively. Based on the results shown in the figure, we draw the following conclusions: (1) for all datasets, the objective function of GPMFS converges rapidly within 20 iterations, indicating the algorithm's efficient convergence capability; (2) Algorithm 1 is effective in optimizing the objective function formulated by GPMFS, demonstrating its practical feasibility. Therefore, we conclude that Algorithm 1 provides a viable solution for obtaining the optimal result.

Next, we will analyze the computational complexity of the PFS-LS algorithm. In the context of multi-label datasets,

where $n \gg L$ and $F \gg L$, the term $L$ can be neglected when calculating the algorithm's complexity. From the workflow of Algorithm 1, it can be deduced that during the preprocessing phase, the complexity of constructing the Pearson correlation matrix $P$ and the Laplacian matrix $L$ is $O(nF^2 + n^2)$. In each iteration, the computational complexity of calculating $A^t$ is $O(nF^2 + F) \approx O(nF^2)$. The complexity of updating $W$ is $O(F^3 + nFL + F^2L) \approx O(F^3 + nF)$, while updating $V$ has a complexity of $O(n^3 + nFL + nL + n^2L) \approx O(n^3 + nF)$. The complexity of updating $U$ is $O(nL) \approx O(n)$. Thus, the total complexity during the iterative phase is $O(n^3 + F^3 + nF^2)$. In the post-processing phase, the complexity of selecting global and personalized features is $O(nF^2)$. Therefore, the overall complexity of the algorithm is $O\big(t(n^3 + F^3 + nF^2) + nF^2 + n^2\big) \approx O\big(t(n^3 + F^3 + nF^2)\big)$, where $t$ denotes the number of iterations.

Finally, we conducted a comparative analysis of the running time across all competing methods. The experiments were conducted on a machine equipped with a 13th Gen Intel(R) Core(TM) i7-13700K CPU and an NVIDIA GeForce RTX 4060 Ti GPU, using Python 3.8 as the programming environment. Table 8 reports the average running time (in seconds) over 10 independent trials on six benchmark datasets(Business, Computers, Education, Enron, Medical and Social) during the feature selection process.

TABLE VIII
Runtime(in second) of competing algorithms across six benchmark datasets.

| Algorithms | FIMF | SCMFS | RFSFS | LSMFS | SLMDS | LCIFS | GPMLS |
|---|---|---|---|---|---|---|---|
| Business | 18.7201 | 0.3119 | 0.0686 | 0.7882 | 3.9266 | 5.8885 | 5.8793 |
| Computers | 35.1948 | 0.6214 | 0.0914 | 1.7052 | 1.8204 | 7.7033 | 7.0467 |
| Education | 28.7593 | 0.4758 | 0.0732 | 1.2979 | 2.9125 | 6.7934 | 7.0117 |
| Enron | 43.9324 | 0.3649 | 0.0663 | 3.4531 | 2.4612 | 7.4411 | 4.0403 |
| Medical | 28.1688 | 1.0903 | 0.0798 | 14.3561 | 6.7565 | 5.7127 | 9.3711 |
| Social | 73.8974 | 0.9797 | 0.1760 | 3.3882 | 2.9135 | 10.3663 | 3.6873 |

As shown in table 8, GPMFS achieves noticeably lower running time than FIMF on larger datasets, and exhibits a comparable computational cost to LCIFS across all datasets. Although the runtime of GPMFS is substantially higher than that of lightweight methods such as RFSFS, this increase primarily stems from the incorporation of matrix inversion operations and the adoption of a personalized feature selection mechanism—both of which are designed to improve the precision of feature evaluation.

Overall, while GPMFS may incur greater computational overhead compared to simpler alternatives, it delivers superior predictive performance by precisely optimizing the feature selection matrix. Thus, the additional cost is justified in practical scenarios. This trade-off between computational efficiency and effectiveness highlights the pivotal role of accurate feature selection in multi-label learning tasks.

## VI. CONCLUSIONS AND FUTURE WORK

In this paper, we propose a novel personalized feature selection method, GPMFS, tailored for multi-label learning scenarios. The method identifies a shared feature set common to all labels, as well as personalized feature subsets specific to each individual label. To construct the shared feature set, we exploit the complex inter-label dependencies by leveraging graph-based representations and label relaxation techniques, thereby facilitating a more informative global selection. For the personalized part, we further utilize the label-specific structure embedded in the feature selection matrix W, and introduce a personalized feature selection threshold to control the extent of label-wise feature selection. Extensive experiments on benchmark datasets under five widely-used evaluation metrics confirm the effectiveness and superiority of our proposed method.

In future research, we will enhance the GPMFS framework by incorporating dimensionality reduction, improving optimization efficiency, and exploring integration with deep learning methods. We also aim to extend the method to dynamic and streaming multi-label environments using adaptive and online learning techniques. These improvements are expected to boost the scalability, robustness, and generalizability of GPMFS across a wide range of tasks.

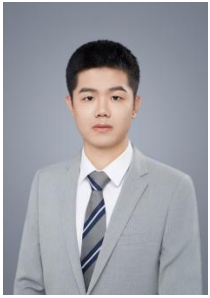
**Yifan Cao** received the B.S. degree in Mathematics and Applied Mathematics from Beihang University,Beijing,P.R.China,in 2022. He is currently pursuing a PhD in Applied Mathematics at the School of Artificial Intelligence, Beihang University, China. His research interests include data science and feature selection.

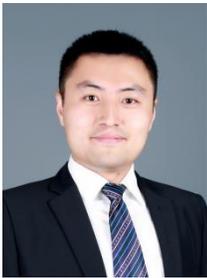
**Zhilong Mi** received the Bachelor's degree in information and computational science and the PhD degree in fundamental mathematics from Beihang University in 2015 and 2021, respectively. He is currently an assistant professor with the Institute of Artificial Intelligence, Beihang University. His main research interests include complex networks, basic theory and security of artificial intelligence.

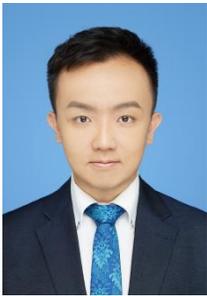
**Ziqiao Yin** received the Bachelor's degree and the PhD degree in fundamental mathematics from Beihang University in 2016 and 2021, respectively. He is a visiting scholar with Yale University between 2019 to 2020. He is currently an assistant professor with the Institute of Artificial Intelligence, Beihang University. His research interests include bionic intelligence and crowd intelligence.

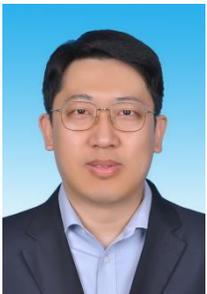
**Binghui Guo** received the Bachelor's degree in applied mathematics and the PhD degree in fundamental mathematics from Beihang University in 2005 and 2010, respectively. He is currently an associate professor with the Institute of Artificial Intelligence, Beihang University. His research interests include data science, complex networks, basic theory and security of artificial intelligence.

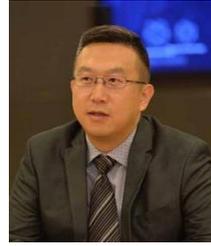
**Jin Dong** is the General Director of Beijing Academy of Blockchain and Edge Computing. He is also the General Director of Beijing Advanced Innovation Center for Future Blockchain and Privacy Computing. The team he led developed "ChainMaker", the first hardware-software integrated blockchain system around the globe. He has been long dedicated in the research areas such as blockchain, artificial intelligence and low-power chip design.